\def\year{2020}\relax
\documentclass[letterpaper]{article}
\usepackage{aaai20}
\usepackage{times}
\usepackage{helvet}
\usepackage{courier}
\usepackage[hyphens]{url}
\usepackage{graphicx}
\usepackage{graphicx}
\usepackage{amsmath}
\usepackage{amssymb}
\usepackage{multirow}
\newcommand{\citealp}[1]{\citeauthor{#1} \citeyear{#1}}

\title{Action Recognition and State Change Prediction in a Recipe Understanding Task \\ 
	Using a Lightweight Neural Network Model \\}
\author{
	Qing Wan,
	Yoonsuck Choe \\
	Department of Computer Science and Engineering, Texas A\&M University\\
	College Station, TX, 77843, USA \\ 
	frankwan@email.tamu.edu, choe@tamu.edu
}

\begin{document}

\maketitle

\begin{abstract}
Consider a natural language sentence describing a specific step in a food recipe. In such instructions, recognizing actions (such as press, bake, etc.) and the resulting changes in the state of the ingredients (shape molded, custard cooked, temperature hot, etc.) is a challenging task. One way to cope with this challenge is to explicitly model a simulator module that applies actions to entities and predicts the resulting outcome (\citealp{bosselut2017simulating}). However, such a model can be unnecessarily complex. In this paper, we propose a simplified neural network model that separates action recognition and state change prediction, while coupling the two through a novel loss function.  This allows learning to indirectly influence each other. Our model, although simpler, achieves higher state change prediction performance (67\% average accuracy for ours vs. 55\% in (\citealp{bosselut2017simulating})) and takes fewer samples to train (10K ours vs. 65K+ by (\citealp{bosselut2017simulating})).
\end{abstract}

\noindent
\section*{Introduction}
Understanding actions in sentences and the resulting changes in the state of the entities is still a challenging task in NLP.  This is due to the various causal factors involved and necessary common-sense knowledge for correct inference of the actions and states. Recently, (\citealp{bosselut2017simulating}) introduced a novel neural process network, specifically designed to understand procedural language with the goal of modeling action dynamics and predicting their effects on entities. However, they had to train their model on a set of over 65K recipes selected from their whole 120K dataset under certain rules. Their proposed model achieved 55\% accuracy on the test set in predicting the state change in the recipe sentences. This is somewhat limited performance, and furthermore, their model was fairly complex, with encoding, action selection, entity selection, state prediction, and action simulation.

To tackle these problems, we introduce a new deep neural network model with a much more concise architecture, and aim at solving a main task that is to correctly recognize actions and predict state changes in a sentence of a recipe. Also, we contribute a novel cost function. Training on any small randomly selected subset (10K samples) of the 65K, our model achieves an average state change prediction accuracy of 67\% on the test set of (\citealp{bosselut2017simulating}), which is an improvement of 12\% over their results.

\section*{Methodology}
Our model is built on  Recurrent Neural Networks (RNN, \citealp{graves2006connectionist}) with Gated Recurrent Units (GRU, \citealp{cho2014emnlp}). In our design, three different types of neural network layers are used: Dynamic RNN (encoder), MLP (decoder), and predictor. We find that this simple architecture is sufficient for action language understanding.

We encode each word in a sentence with a one-hot vector/code \cite{hinton1984distributed} and count a whole sequence of vectors of words that are generated by encoding to produce the input representation. That is, in our model, we do not use any word embedding.

We use two different metrics, loss function and error function, for training control. We train our model by loss function and validate the model by error function.

We design the loss function under the scope of tangent. Let $\mathbf{Y},\mathbf{P}\in[0,1]^m$ (finite dimensional space) and $\mathbf{Y}$ is a label, $\mathbf{P}$ is a prediction for $\mathbf{Y}$, then the loss function is defined as
\begin{equation} l(\mathbf{Y},\mathbf{P})=\sum_{i=1}^{i=m}10\times\tan(\frac{\pi}{2}|\mathbf{Y}(i)- \mathbf{P}(i)|).
\end{equation}
Then, $l(\mathbf{Y},\mathbf{P})=0$ iff $\mathbf{Y}=\mathbf{P}$ due to the nonnegativity of $\tan((\frac{\pi}{2}|\mathbf{X}|)$. With the help of the following property
\begin{equation}
\tan(|x|)\ge|x|,x\in(-\frac{\pi}{2},\frac{\pi}{2}),
\end{equation}
preservation of triangle inequality and convexity can be verified.

We define an error function with the help of cross entropy and the error is
expected to serve as an accuracy indicator that will decide when to save our model during training. Cross entropy for discrete topology on $\mathcal{X}$ is well-defined as:
\begin{equation}
H(p,q)=-\sum_{x\in\mathcal{X}} p(x)\log_2 q(x).
\end{equation}
Our error function $\epsilon(\mathbf{Y},\mathbf{P})$ is defined by
\begin{equation}
\epsilon(\mathbf{Y},\mathbf{P})=|H(P_{\mathbf{P}},Q_{\mathbf{Y}} )-H(Q_{\mathbf{Y}}, Q_{\mathbf{Y}} )|,
\end{equation}
where $P_{\mathbf{P}}$ and $Q_{\mathbf{Y}}$ denotes probability mass function (pmf) generated by the predicted vector $\mathbf{P}$ and the label $\mathbf{Y}$, respectively; and a Softmax function can implement the generation of $P_{\mathbf{P}}$ and $Q_{\mathbf{Y}}$. Qualitatively, the more accurately a neural network predicted, a lower $\epsilon$ should  be achieved. This is guaranteed by
\begin{eqnarray}
&&	|H(P_\mathbf{P},Q_\mathbf{Y})-H(Q_\mathbf{Y},Q_\mathbf{Y})|  \nonumber \\
&&	\le \sum_{x \in \mathcal{X}} |(Q_{\mathbf{Y}}(x) - P_{\mathbf{p}}(x))\log_2 Q_{\mathbf{Y}}(x)|
\nonumber \\
&&	\le \max_{x \in \mathcal{X}} \{|\log_2 Q_{\mathbf{Y}}|\} \sum_{x \in \mathcal{X}} |Q_{\mathbf{Y}}-P_{\mathbf{P}}| \nonumber,
\end{eqnarray}
where the last $\le$ is due to bounded $|\log_2 Q_{\mathbf{Y}}|$ for finite dimensions. So a continuity will follow. This continuity warrants the use of it as an accuracy indicator for model saving.

\section*{Experiments}
This section describes how the experiments are conducted.

\subsection*{Data Source \& Data Selection}

We used the dataset and corpora from (\citealp{bosselut2017simulating}). It has a total of 120K samples. Bosselut et al. used carefully filtered 65,815 samples from the whole 120k for training, and 693 samples for testing. Besides, they provided corpora (token libraries) for vocabularies of sentences(text), verbs and state changes.

In our experiment, we randomly selected a 10K subset (15\%) from their 65,815 samples and separate this 10k into 9k for training loss function and 1k for validating error function. We use their corpora for text, verb and state change. We add ``UNK'' (unknown) type to each corpus for vocabulary out of their corpora.

\subsection*{Encoding/Decoding}
We directly feed a one-hot word vector (no embedding) into a GRU cell each time. We pad each sentence to the maximum length of sentences in the recipe. Our model has two encoding layers each with 1600 units and 800 units, respectively. Both verb decoding and state change decoding are analyzed by an independent MLP with 500 hidden units. Each MLP is connected to the final state of the GRU encoding layer.

\subsection*{Loss Function}
To ovecome finite word-length effect, a modified loss,
\begin{equation}
l(\mathbf{Y},\mathbf{P})=\sum_{i=1}^{i=m}10\times\tan(0.499\pi\times|\mathbf{Y}(i)-
\mathbf{P}(i)|),
\end{equation}
is applied in practice and turn $l(\mathbf{Y}, \mathbf{P})$ into a bounded function.

Two loss functions, action loss and state loss, are defined and are summed to produce total loss, based on which learning is conducted.

\subsection*{Training, Validating, Testing}
Training is conducted on a GPU with the RMSProp optimizer \cite{tieleman2012lecture} under a learning rate of 0.0001. We train total loss on 9K samples for 201 epochs within which we validate error function on 1K samples every two epochs. If a lower error monitored during training, the model would be stored.

After training finished, the trained model will be tested on (\citealp{bosselut2017simulating})'s 693 test set. One missing tolerance, compatible with their benchmark, is applied. It allows a prediction to have one missing item at most when compared with the label. Accuracy is measured as percentage of sentence predictions satisfying the rule.

\section*{Results}
We independently repeat our experiments for 4 times and report accuracies on both action recognition and state change predition (Table 1). Note that (\citealp{bosselut2017simulating}) does not report the action accuracy.

\begin{table}[h]
	\caption{Test our model on Bosselut's 693 test set}
	\label{Table1}
	\resizebox{.47\textwidth}{!}{\begin{tabular}{|c|c|c|c|c|}  
		\hline
		Our experiment No. &1 &2 &3 &4\\
		\hline
		Action acc (\%) &81.2 &80.9 &80.9 &80.9  \\
		\hline
		State acc (\%)& \textbf{66.6} &\textbf{66.6} &\textbf{67.0} &\textbf{67.2}  \\
		\hline
		\multirow{2}{*}{Bosselut's Model}&Action acc (\%) &\multicolumn{3}{c|}{N/A} \\
		\cline{2-5} & State acc (\%) &\multicolumn{3}{c|}{55} \\
		\hline
	\end{tabular}}
\end{table}

Our results significantly outperform those in (\citealp{bosselut2017simulating}) under the same benchmark while using only a 15\% subset of data for training.

\section*{Conclusion}
We contribute a lightweight neural network with a novel cost function. With an improvement of 12\% and significantly smaller training data used, our model outperforms a previous language understanding model where the goal is to recognize verbs and predict state changes.
\small
\bibliography{SA-WanQing}
\bibliographystyle{aaai}

\end{document}